\documentclass[letterpaper, 10 pt, conference]{ieeeconf}  %

\IEEEoverridecommandlockouts                              %

\usepackage{graphics} 
\usepackage{epsfig} 
\usepackage{mathptmx} 
\usepackage{times} 
\usepackage{amsmath} 
\usepackage{amssymb}  

\usepackage{graphicx}
\usepackage{subcaption}
\usepackage{cite}
\usepackage{bm}
\usepackage{xfrac}
\usepackage[T1]{fontenc}

\title{\LARGE \bf
A Ducted Fan UAV for Safe Aerial Grabbing and Transfer of Multiple Loads Using Electromagnets
}

\author{
    Zhong Yin$^{1}$ and Hailong Pei$^{2}$ 
    \thanks{
    All authors are with the Key Laboratory of Autonomous Systems and Networked Control, Ministry of Education, Guangdong Engineering Technology Research Center of Unmanned Aerial Vehicle Systems, South China University of Technology, Guangzhou,510640, China
    \{202110183002\}@mail.scut.edu.cn,\{auhlpei\}@scut.edu.cn.
    }
}

\begin{document}
\maketitle
\thispagestyle{empty}
\pagestyle{empty}

\begin{abstract}
In recent years, research on aerial grasping, manipulation, and transportation of objects has garnered significant attention. 
These tasks often require UAVs to operate safely close to environments or objects and to efficiently grasp payloads.
However, current widely adopted flying platforms pose safety hazards: unprotected high-speed rotating propellers can cause harm to the surroundings.
Additionally, the space for carrying payloads on the fuselage is limited, and the restricted position of the payload also hinders efficient grasping.
To address these issues, this paper presents a coaxial ducted fan UAV which is equipped with electromagnets mounted externally on the fuselage, enabling safe grasping and transfer of multiple loads in midair without complex additional actuators. 
It also has the capability to achieve direct 
human-UAV cargo transfer in the air.
The forces acting on the loads during magnetic attachment and their influencing factors were analyzed. 
An ADRC controller is utilized to counteract disturbances during grasping and achieve attitude control. 
Finally, flight tests are conducted to verify the UAV's ability to directly grasp multiple loads from human hands in flight while maintaining attitude tracking.
\end{abstract}

\section{Introduction}
In recent years, with the increasing maturity of control technology and the rise of hardware reliability, operations interacting with the environment such as aerial grasping or load transportation have gradually attracted widespread attention. 
For example, some studies have enabled drones to grasp and transport payloads by attaching simple gripping mechanisms underneath the fuselage \cite{c1} \cite{c41}. However, this limits the drone to approaching the load only from directly above, which restricts its practicality.
Another approach is to install robotic arms on the UAV body that can be used for grasping and manipulation, such as \cite{c2,c4,c6,c7}.  or the combination of drones.
During aerial manipulations, the aircraft is supposed to be close to the target object and to perform flight tasks near surroundings. 
However, as detailed in reviews \cite{c10,c11,c12}, flying platforms currently used in research are mainly multirotors and helicopters, which have unprotected high-speed rotating blades that can cause serious damage.
For this reason, there are studies focusing on the protective mechanisms of drones \cite{c15}\cite{c16}. Although these structures provide some protection, they do not contribute lift but are purely "dead weight" which limits flight endurance.

\begin{figure}
    \centering
    \includegraphics[width=0.8\linewidth,]{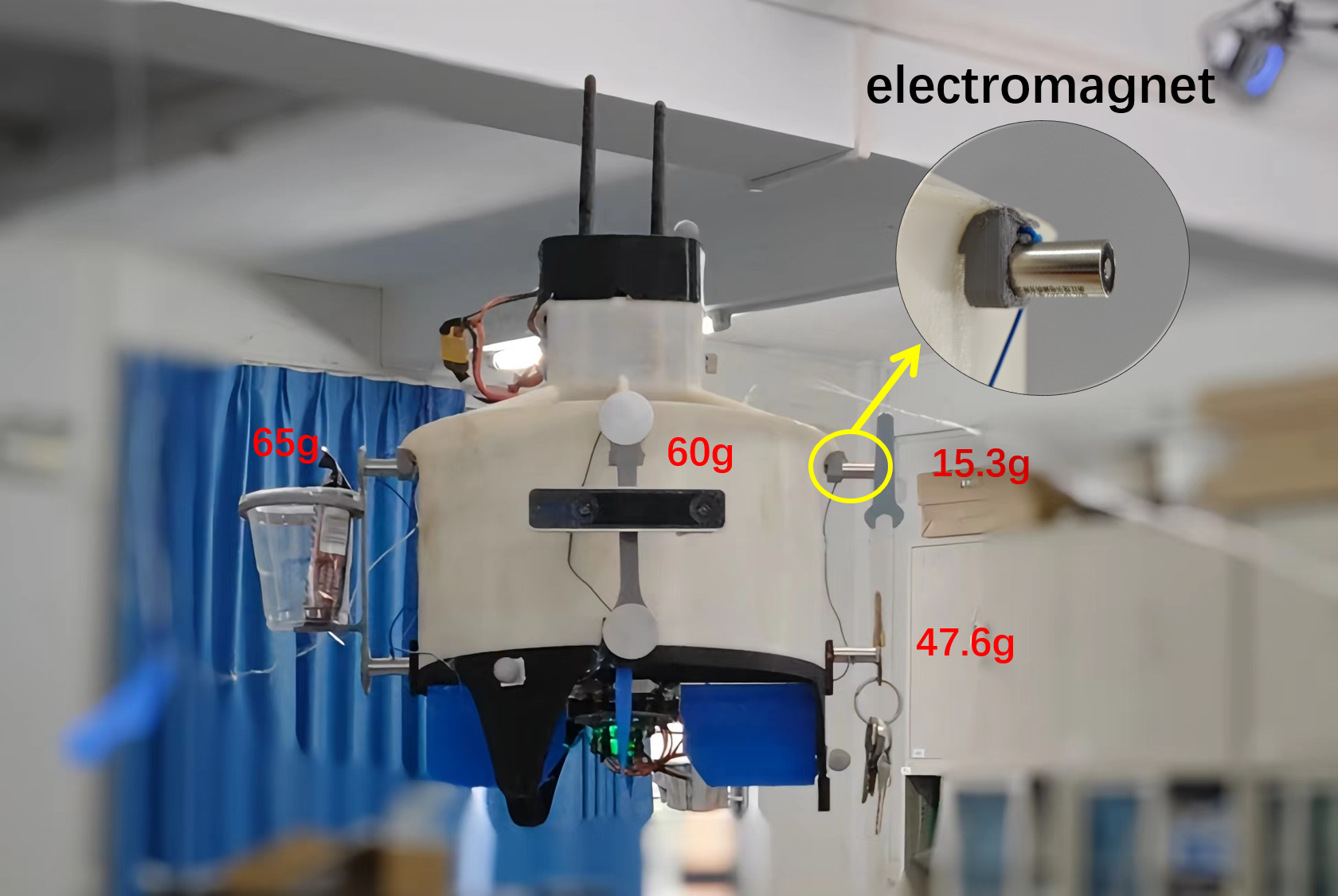}
    \caption{The novel design of DFUAV called Ductopus. It's a coaxial ducted fan UAV with eight electromagnets(like an octopus with eight tentacles), which enables attaching and detaching multiple payloads safely in mid-air.}
    \label{fig:0}
\end{figure}

In the field of unmanned aerial vehicles, ducted fan UAVs (DFUAVs) have garnered significant attention in recent years. A distinctive feature of these UAVs is that their moving and actuating parts are shielded by an annular fuselage known as the duct. This design eliminates damage caused by propeller scratches.
Moreover, this configuration allows the propeller to generate greater hovering thrust compared to an isolated one of the same diameter, owing to the shaped duct \cite{c17}.

These characteristics make it promising for interactive tasks such as grasping and contact inspections.
In \cite{c18}, the robust trajectory tracking in the scenario of contact between a DFUAV and a wall is discussed, but only simulations were conducted without practical experiments. In \cite{c19}, inspection-by-contact using a DFUAV with a robotic arm is studied. In \cite{c33} and \cite{c40}, the coaxial ducted fan UAV design is preferred for various applications due to its cancellation of motor reaction torque and reliable structure.
However, as will be illustrated later in Sec. II, while DFUAV's structure is compact and efficient, its space for payload deployment is confined. 
This drawback limits their availability for carrying payloads and performing more aerial manipulation tasks.

To address these issues, in this paper, we introduce a newly designed coaxial ducted fan UAV called the Ductopus, which has eight electromagnet-based attachment points(like an octopus with eight tentacles) and is able to attach multiple payloads in mid-air safely and conveniently without the need for additional actuators, addressing the issue of limited payload space in previous DFUAVs while retaining their advantages like compactness and efficiency.

This new design faces several control challenges, such as long-lasting disturbance moments caused by loads, for which we employ a linear adaptive disturbance rejection controller (LADRC) which uses an extended state observer (ESO) to observe the collective effects of external perturbations and model changes in real-time and compensates them into the controller \cite{c38}.
 
A series of indoor flight experiments were conducted and verified the feasibility of the proposed design and control algorithms, allowing individuals to safely approach the flying Ductopus and directly attach various payloads to it.

\begin{figure}[t]
    \centering
    \subcaptionbox{}{\includegraphics[width=0.34\linewidth, height=0.75in]{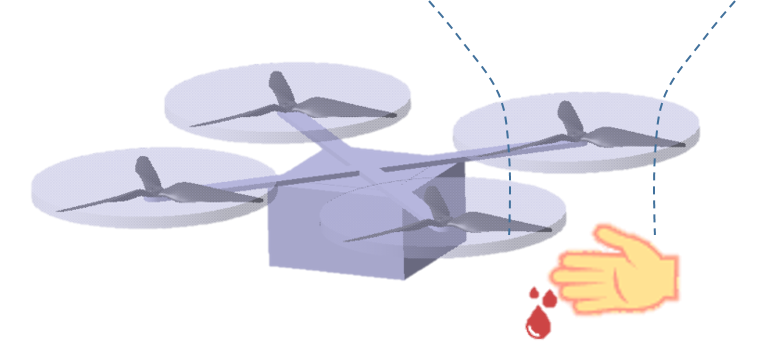}}
    \subcaptionbox{}{\includegraphics[width=0.28\linewidth, height=0.75in]{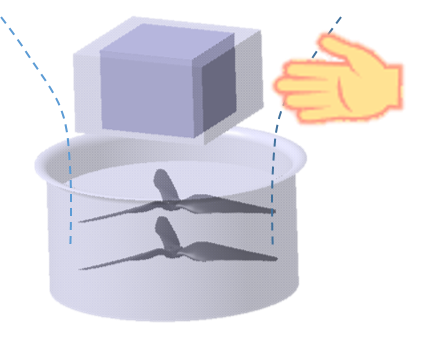}}
    \subcaptionbox{}{\includegraphics[width=0.3\linewidth, height=0.75in]{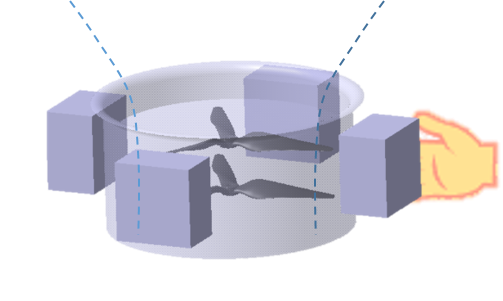}}
    \caption{Illustrations of UAVs carrying loads and interacting with individuals. The dotted line indicates the airflow. (a)Multirotors: The position of the payload is inconvenient for loading and unloading, and the rotors may cause scratches.(b)DFUAV: Confined space for payloads and dangerous to reach. (c)The new design proposed in this paper facilitates the safe loading and unloading of multiple payloads.}
    \label{fig:0}
\end{figure}
 
The rest of this paper is structured as follows: In Section II, we consider the pros and cons of common practical configurations for aerial manipulations.
And outlines the structure design of the UAV presented in this paper. Section III establishes a dynamic model of this system. 
Section IV analyzes the forces during the attachment of loads.
Section V achieves attitude control for the proposed UAV using LADRC. Section VI details the conducted
experiments, including parameter identifications and
indoor flight experiments.

\section{Comparisons and System Design}
\begin{figure}[]
    \centering
    \begin{subfigure}[]{0.45\columnwidth}
        \includegraphics[width=1.2\linewidth, height=1.8in]{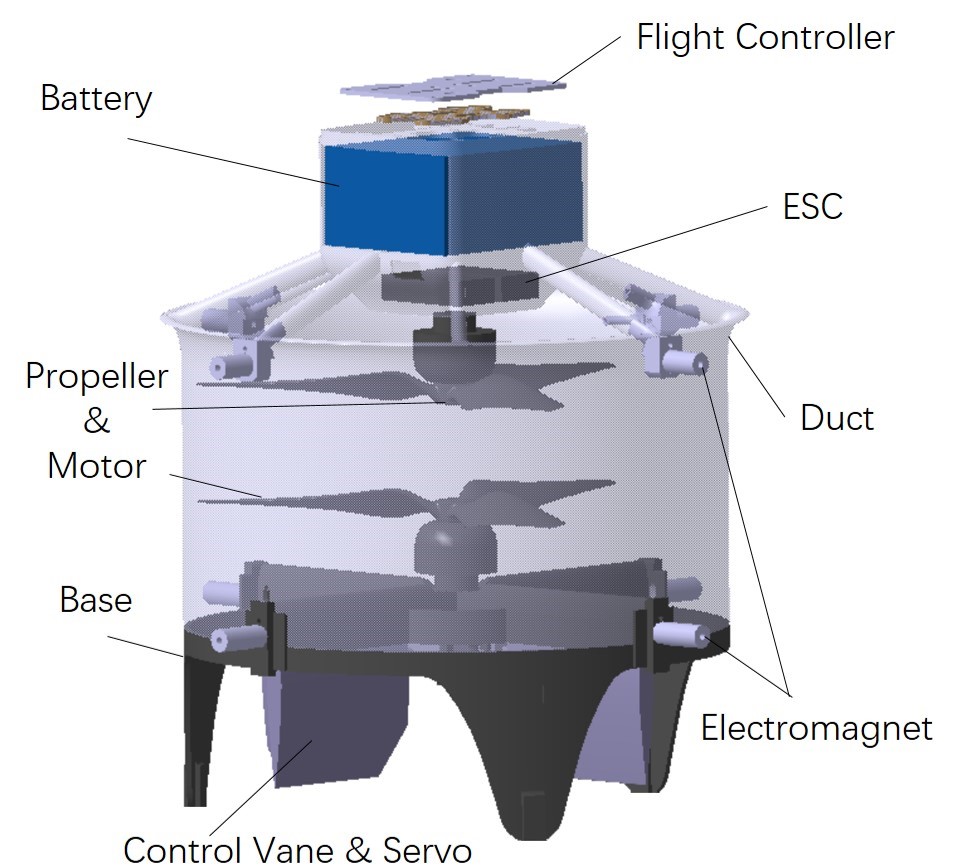}
        \caption{}
        \label{fig:1a}
    \end{subfigure}
    \hfill
    \begin{subfigure}[]{0.45\columnwidth}
        \includegraphics[width=\linewidth, height=1.8in]{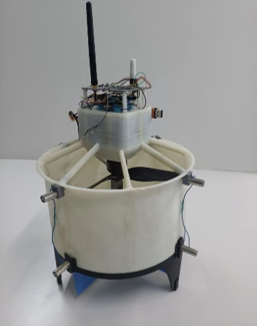}
        \caption{}
        \label{fig:1b}
    \end{subfigure}
    \caption{(a)The structure and main components of the Ductopus. (b)The prototype used in this paper.}
    \label{fig:1}
\end{figure}

Whether dedicated to transporting cargo or carrying various functional payloads, sufficient payload space and convenient loading and unloading methods have always been essential to consider in drone design. At the outset of our work, we compared the payload space limitations among common types of existing UAVs. 
As depicted in Fig.\ref{fig:0}, for the typical multirotor UAVs, constrained by their inherently underactuated nature, payload positioning is often restricted to the center of the fuselage, which is detrimental to payload installation and unloading. 
In the case of helicopters, larger rotor blades, while providing greater efficiency, also bring more severe potential damages, making aerial payload handling similarly impractical. For existing DFUAV designs, as summarized in \cite{c17}, payloads can only be placed at the top center of the fuselage for balance considerations. However, this location coincides with the intake of the duct, and an increase in payload volume significantly affects the airflow entering the duct. Additionally, the strong airflow suction at the duct inlet poses safety hazards for loading and unloading payloads during flight.
To address these issues, the aircraft designed in this paper innovatively adopts an external loading method on the duct. The Ductopus's main structure, as depicted in Fig.\ref{fig:1}, mainly consists of a duct and a support base made of ABS material through 3D printing. The upper part of the duct houses a 10000mAh battery, an STM32-based flight control board, and communication components. Inside the duct, a pair of brushless motors (iFlight XING 2814 1100kv) are coaxially mounted, equipped with ESCs (Hobbywing Platinum 25A) and counter-rotating nine-inch three-blade propellers (iFlight HQProp 9x5x3). Four servos (KST X08H) are fixed on the base with rotatable control surfaces. The selected cylindrical electromagnet (CNYRIL P10/25) has a diameter of 10mm and a height of 25mm, with screw holes for direct attachment to the fuselage. Its nominal holding force under 12V voltage is 2 kgf/0mm. 

\begin{figure}[b]
    \centering
    \begin{subfigure}[]{0.45\columnwidth}
        \centering
        \includegraphics[width=0.75\linewidth]{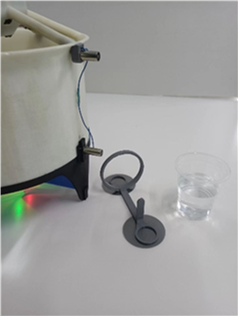}
        \hspace{0.001\columnwidth} %
        \caption{}
        \label{fig:2a}
    \end{subfigure}
    \begin{subfigure}[]{0.45\columnwidth}
        \centering
        \includegraphics[width=0.75\linewidth]{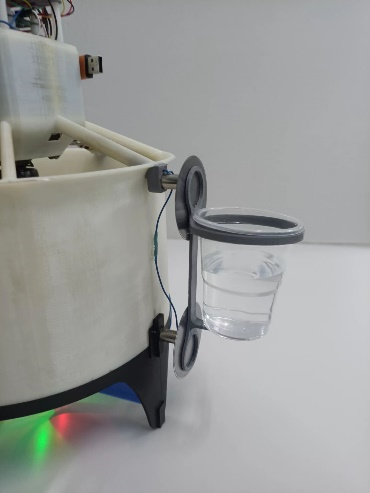}
        \hspace{0.001\columnwidth} %
        \caption{}
        \label{fig:2b}
    \end{subfigure}
    \caption{Combining electromagnets with specialized load modules facilitates various load attachments. (a) A plastic cup, filled with water as load, is mounted using 3D printed modules. (b) Magnetic attachment facilitates the attachment of loads.}
    \label{fig:2}
\end{figure}
During flight, the flight controller receives attitude feedback signals from the Optitrack motion capture system and runs control algorithms at a cycle time of 5ms. 
Additionally, it controls the electromagnets' magnetization by switching relays using PWM signals. Two sets of propellers rotate in opposite directions, providing the main lift and enabling yaw control through speed differential.
Four groups of servos at the bottom drive the control vanes to generate attitude control torque \cite{c31}.
The total weight of the aircraft is 1.59kg (including a battery weighing 0.6kg). Each of the eight electromagnets can attract and hold magnetic objects weighing approximately 60 grams.
To carry non-magnetic loads, a mounting module weighing 15g was designed to contain a payload, consisting of 3D printed parts, a magnetic attractor plate, and a cup. Fig. \ref{fig:2} illustrates the attachment of a payload (a cup of water weighing 90g) using
electromagnets. 
 
\section{System Modeling}
\begin{figure}[b]
        \centering
        \includegraphics[width=0.5\linewidth,]{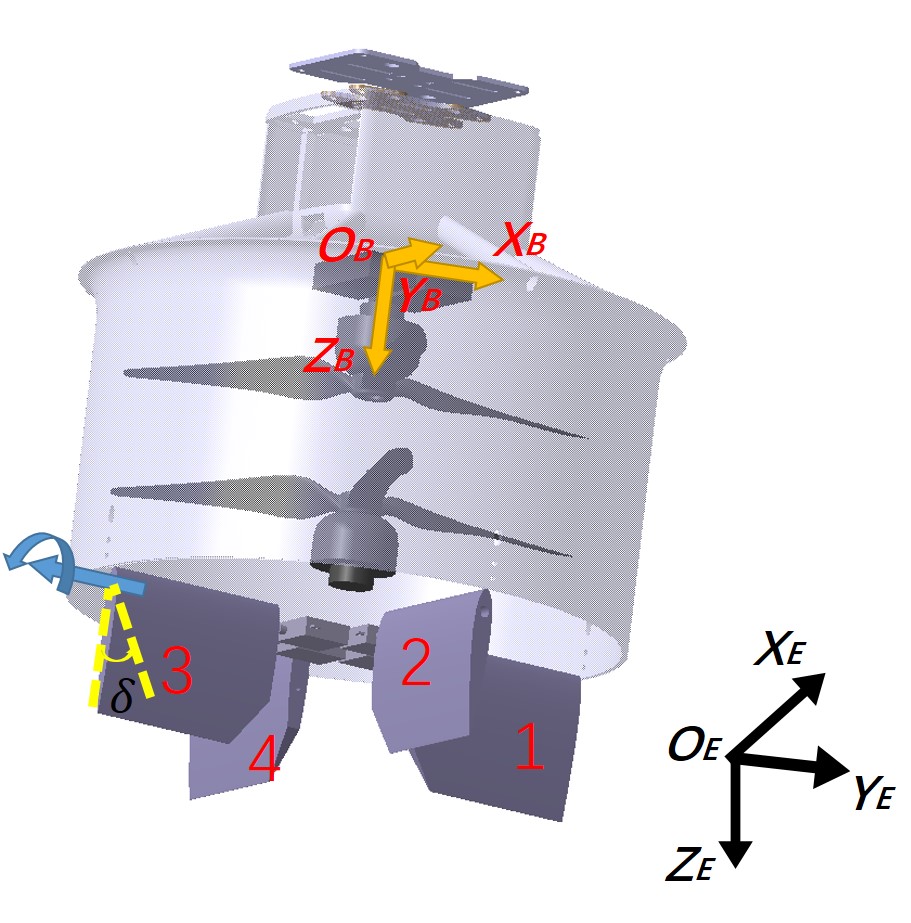}
        \label{fig:3a}
    \caption{Coordinate frames of the Ductopus and the sequence of servos.}
    \label{fig:3}
\end{figure}

The aerodynamic forces acting on a DFUAV are intricate, exhibiting nonlinear and asymmetric characteristics that vary with the tilt of the aircraft \cite{c29}.
Moreover, when loads of unknown weight are attached during flight, the parameters of the vehicle model will change, including total weight and inertia.  
However, the overall impact of these factors will be observed by ESO, which allows these issues to be ignored during modeling process in this section, resulting in a reasonably simple model.  
Specifically, consider the forces and moments on the system are mainly generated by two components, the propellers and the control vanes.   
 
As depicted in Fig.\ref{fig:3}, the coordinate system required for system modeling is defined as follows: the inertial frame is represented as $\left\{
    {{O_E} - {X_E}{Y_E}{Z_E}} \right\}$ which is in the convention of
North-East-Down (NED).The origin of the aircraft body frame $\left\{ {{O_B} -
    {X_B}{Y_B}{Z_B}} \right\}$ is located at the center of gravity of the drone.
The z-axis is parallel to the motor axis and points towards the ground. Euler angles $\eta = {\left[ {\begin{array}{*{20}{c}}\varphi &\theta &\psi \end{array}} \right]^T}$are used to represent the aircraft attitude, where roll, pitch, and yaw angles are respectively represented as $\varphi$,$\theta $\ and $\psi$. 

\subsection{Propeller force and torque} 
The forces on the duct are mainly generated by the rotation of the two propellers. According to \cite{c30}, the resultant force on the duct $\mathbf{F}_{\text{prop}}$ can be effectively modeled as: 
\begin{equation}
    \label{DuctForce}
    \mathbf{F}_{\text{prop}} = \begin{bmatrix}
        0         & 0         \\
        0         & 0         \\
        C_{Tz\_1} & C_{Tz\_2}
    \end{bmatrix}
    \begin{bmatrix}
        \Omega _1^2 \\
        \Omega _2^2
    \end{bmatrix},
\end{equation}

where ${C_{Tz\_1}}$,${C_{Tz\_2}}$ are propeller parameters and their values are obtained experimentally in section VI. $\Omega _1$ and $\Omega _2$ represent the rotational speeds of the upper and lower motors respectively. 
Since the design of contra-rotation coaxial rotors can counteract most of the gyroscopic torque, the torque of the propellers can be modeled as: 
\begin{equation}
    \label{ductmoment}
    \mathbf{M}_{\text{prop}} = \begin{bmatrix}
        0         & 0         \\
        0         & 0         \\
        C_{Mz\_1} & C_{Mz\_2}
    \end{bmatrix}
    \begin{bmatrix}
        \Omega _1^2 \\
        \Omega _2^2
    \end{bmatrix},
\end{equation}
where $C_{Mz\_1},_{Mz\_2}$ are fixed parameters relative to propellers. 
\subsection{Vane force and torque} 
The sequence of the servos is illustrated in Fig. \ref{fig:3}, with the rotational angles of the servos defined as ${\delta } = {\left[ {\begin{array}{*{20}{c}}
    {{\delta _1}}&{{\delta _2}}&{{\delta _3}}&{{\delta _4}}
    \end{array}} \right]^T}$.  
They control the deflection of vanes, changing the airflow at the outlet of the duct, and the resulting aerodynamic force induced on the i-th control surface can be modeled as \cite{c31}:
\begin{equation}
    \label{vaneforece}
    {F_{vane\_i}} = \frac{1}{2}\rho {A_{cv}}{C_{Lcv}}V_e^2{\delta _i},
\end{equation}
where $\rho$ is the density of the air, ${A_{cv}}$ is the reference aerodynamic area, ${V_e^2}$ is the air velocity at the duct outlet and ${C_{Lcv}}$ is the slope of the lift line of the vane. 
This force is typically negligible compared to the rotor thrust. However, the torque it induces is crucial for attitude control. Based on the symmetric design of the control vanes, the resultant moment in the body-fixed frame can be modeled as: 
\begin{equation}
    \label{vanemoment}
    \mathbf{M}_{\text{vane}} = \frac{1}{2}\rho A_{cv} C_{Lcv} V_e^2 \begin{bmatrix}
        -l_1 & 0    & l_1  & 0 \\
        0    & -l_1 & 0    & l_1   \\
        l_2  & l_2  & l_2  & l_2       
    \end{bmatrix}
    \begin{bmatrix}
        \delta_1 \\
        \delta_2 \\
        \delta_3 \\
        \delta_4
    \end{bmatrix},
\end{equation}
where ${l_1}$ and ${l_2}$ represent the distances from the aerodynamic point of the control vane to x-axis and z-axis in body reference, respectively, and ${V_{e}}$ is the outlet air speed of the duct. 
\subsection{Total Forces and Torques}
In conclusion, the total force and torque acting on the UAV body frame are
modeled as two components:
\begin{equation}
    \label{totalforce}
    \begin{aligned}
        \mathbf{F} & = \mathbf{F}_{\text{grav}} + \mathbf{F}_{\text{prop}},
    \end{aligned}
\end{equation}
\begin{equation}
    \label{totalmoment}
    \begin{aligned}
        \mathbf{M} & = \mathbf{M}_{\text{prop}} + \mathbf{M}_{\text{vane}},
    \end{aligned}
\end{equation}
where ${\mathbf{F}_{\text{grav}} = m{\mathbf{R}_{\text{gb}}}\mathbf{G}}$ is the
gravitational force expressed in the body frame. $m$ denotes the total mass of
the aircraft and ${\mathbf{R}_{\text{gb}}}$ represent the rotation matrix from
the inertial frame to the body frame following the Z-Y-X order. $\mathbf{G}$ is
the gravitational acceleration vector in the inertia frame.
\subsection{Attitude dynamics}
UAV flying dynamic follows the Newton-Euler equations and in the body
coordinate system, there are:
\begin{equation}
    \label{Newton-Euler equations}
    \begin{aligned}
        \mathbf{J} \dot{\omega} + (\omega \times \mathbf{J} \omega) & = \mathbf{M},
    \end{aligned}
\end{equation}
\begin{equation}
    \label{attitude dynamic}
    \dot{\eta} = \mathbf{Q}\omega, \\
\end{equation}
where $\omega = {\left[ {\begin{array}{*{20}{c}}
                    p & q & r
                \end{array}} \right]^T}$  
represents the angular velocity of the aircraft, ${\mathbf{J}\in {\mathbb{R}^{3 \times 3}}}$ represents the inertia matrix of the aircraft body, in this paper, due to the symmetrical structure of the aircraft, we assume ${\mathbf{J}}$ a diagonal matrix. $\mathbf{Q}$ represents the conversion between angular velocity and Euler angle change rate.
\subsection{Control allocation}
It should be noted that equation \eqref{vanemoment} is underdetermined, meaning multiple solutions of servo angle ${\delta }$ may exist for a given desired control moment \cite{c36}. Similarly, in \eqref{totalmoment}, both motor speed and servo angle can contribute to the torque of the z-axis. 
This provides the aircraft with additional fault tolerance.
In this paper, to eliminate the redundancy in actuation, we choose to control yaw angle solely through motor speed differential, while the servos are used to control pitch and roll angle. And let the opposite sets of servos always have equal and reverse deflection
angles $\Delta {\delta _x} = - {\delta _1} = {\delta _3},\Delta {\delta _y} =
        - {\delta _2} = {\delta _4}$. This leads
        to the same deflection of vanes in the body frame. 
        So given a desired torque for the three axes $\left[ {\begin{array}{*{20}{c}}
    {{\tau _x}}&{{\tau _y}}&{{\tau _z}}
    \end{array}} \right]$, according to \eqref{vanemoment},the allocation of actuators will be:
\begin{equation}
    \label{torque allocation}
    \begin{array}{l}
        {\tau _x} = 0.5\rho {A_{cv}}{C_{Lcv}}V_e^2\left( { - {\delta _1} + {\delta _3}} \right) = {C_{{M_\delta }}} \cdot \Delta {\delta _x},\\
        {\tau _y} = 0.5\rho {A_{cv}}{C_{Lcv}}V_e^2\left( { - {\delta _2} + {\delta _4}} \right) = {C_{{M_\delta }}} \cdot \Delta {\delta _y},\\
        {\tau _z} = {C_{Mz\_1}} \cdot \Omega _1^2 + {C_{Mz\_2}} \cdot \Omega _2^2,
    \end{array}
\end{equation}
where ${C_{{M_\delta }}}$ represents the control effectiveness coefficient. And from \eqref{DuctForce}, the desired thrust $T_d$ is related to the speed of the two motors:
\begin{equation}
    \label{thrust}
    \begin{array}{l}
        {T_d} = {C_{Tz\_1}}\Omega _1^2 + {C_{Tz\_2}}\Omega _2^2.
    \end{array}
\end{equation}
By utilizing \eqref{torque allocation} and \eqref{thrust}, the motor speed $\left[ {\begin{array}{*{20}{c}}
    {{\Omega _1}}&{{\Omega _2}}
    \end{array}} \right]$ and servo angles ${\bf{\delta }}$ can be uniquely determined and rapidly computed for a given desired torque and thrust.
 
\begin{figure}[b]
    \centering
    \includegraphics[width=0.7\linewidth,]{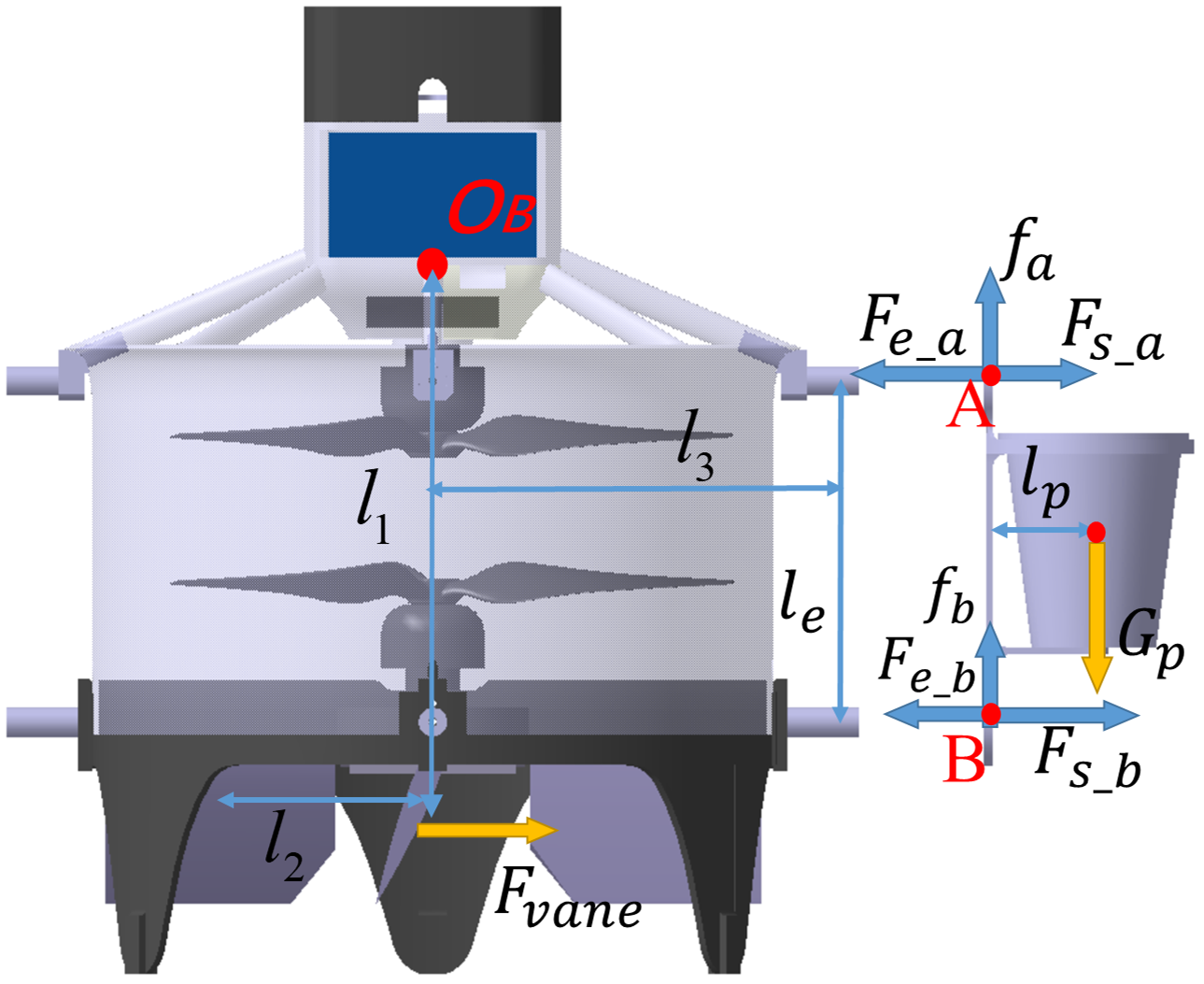}
\caption{Force diagram of a unilateral attached load. (For clarity, a distance is maintained between the load and the contact point)}
\label{fig:6_2}
\end{figure}
\section{Load Attachment Analysis}
To ensure the load can be firmly attached and the UAV's attitude remains controllable, certain factors are considered in this section. 
\subsection{Conditions for stable attachment}
For example, as shown in Fig.\ref{fig:6_2}, when two mounting points are used to attach the load, assuming the points of contact between the load and the UAV are A and B respectively, and the aircraft is in a stable hovering state, analyzing the forces acting on the load separately, we have: 

In the vertical direction, the friction force $f_a$ and $f_b$ at the contact points need to balance the gravity force of the load $G_p$:
\begin{equation}
    \label{1}
    \begin{aligned}
        {f_a} + {f_b} = {G_p}.
    \end{aligned}
\end{equation}

In the horizontal direction, the contact points experience magnetic attraction forces ${F_{e\_a}}$, ${F_{e\_b}}$ and supporting forces ${F_{s\_a}}$, ${F_{s\_b}}$, and are in force equilibrium:
\begin{equation}
    \label{2}
    \begin{aligned}
        {F_{e\_a}} + {F_{e\_b}} = {F_{s\_a}} + {F_{s\_b}}.
    \end{aligned}
\end{equation}

Taking point B as the pivot, the torque equilibrium is:
\begin{equation}
    \label{3}
    \begin{aligned}
        \left( {{F_{e\_a}} - {F_{s\_a}}} \right){l_e} = {G_p}{l_p},
    \end{aligned}
\end{equation}
where ${l_e}$ is the distance between points A and B. ${l_p}$ represents the moment arm from the gravity of the load to the pivot.
The maximum static frictional force before slippage occurs at the contact points is determined by:
\begin{equation}
    \label{4}
    \begin{aligned}
        {f_a} = {k_f}{F_{s\_a}},{f_b} = {k_f}{F_{s\_b}},
    \end{aligned}
\end{equation}
where ${k_f}$ is the coefficient of friction of the contact surface.
To not drop the load when attaching it, taking contact point A as an example, the following conditions need to be satisfied:
Vertically, the friction force exceeds the threshold value:${f_a} > {G_{p\_a}}$, where the threshold value $0 < {G_{p\_a}} < {G_p}$.
Horizontally, support force remain positive: ${F_{s\_a}} > 0$ 

In our experiment, when the load becomes heavier, point A will loosen first causing the load to fall off, and vertical loosening tends to occur before horizontal loosening. From \eqref{3} and \eqref{4}, it can be obtained:
\begin{equation}
    \label{5}
    \begin{aligned}
        {f_a} = {k_f}\left( {{F_{e\_a}} - {G_p}\frac{{{l_p}}}{{{l_e}}}} \right).
    \end{aligned}
\end{equation}

It can be seen that, to enhance load attaching capability, besides increasing the electromagnet attraction force ${F_{e\_a}}$, increasing $l_e$, reducing $l_p$, and increasing the friction coefficient $k_f$ of the contact surface would all be helpful. In our tests, a single electromagnet can attract and maintain a weight of approximately 60g. Two electromagnets on the same side can jointly bear a load of about 110g.

\subsection{Influence of load}
The attachment of load exerts a gravitational force on the fuselage as well as its resulting moment. 
The gravity is overcome by the lift of the propellers, which is often given sufficient margin during aircraft design.
Our work is more concerned with attitude control.
The torque caused by the load will induce aircraft rotation, which needs to be balanced by control vane deflection. 
For example, in the case of Fig.\ref{fig:6_2}, a load located in the positive y-axis of the aircraft will cause the aircraft to rotate around the center of gravity ${O_B}$, requiring aerodynamic force ${F_{vane}}$ generated by control vanes along the x-axis to balance it. Combining \eqref{torque allocation}, there is:

\begin{equation}
    \label{6}
    \begin{aligned}
        {C_{{M_\delta }}} \cdot \Delta {\delta _x} = {G_p}\left( {{l_p} + {l_3}} \right),
    \end{aligned}
\end{equation}
where ${l_3}$ is the distance from the contact point to the z-axis of the fuselage. 
Due to mechanical constraints, the servo's deflection angle 
$\Delta {\delta_x}$ is limited, when it deflects to balance the load, the available rotation space decreases, reducing the torque available for attitude control. Therefore, the maximum control moment provided by the vanes also limits the load capacity. When attaching multiple loads, it is preferable to distribute the load weight symmetrically to reduce the torque interference caused by the load.
In Sec. VI, it was tested that the maximum torque that the control vanes can provide at near hovering is 0.12Nm. 
By employing \eqref{6}, the prototype in this paper was calculated to be capable of balancing approximately 70g of unilateral load.
\section{Attitude Control}
As indicated in \cite{c11}, aerial interactions often pose challenges to the control of aircraft. In our case, when the vehicle is attaching payloads mid-air, the gravity of the load does not align with the center of gravity of the aircraft, resulting in sudden, unknown, and long-lasting disturbances to the vehicle. Moreover, the total mass and inertia of the system changes, which may also cause a deterioration in control effectiveness.
To address these issues, we employ a Active Disturbance Rejection Controller (ADRC) for attitude control. 
The ADRC method, introduced by Han in the 1990s \cite{c22} \cite{c23}, tackles unmodeled features and external interferences by treating them as total disturbances and incorporating them into the system's extended state. Real-time observation of the extended state is achieved through an Extended State Observer (ESO), enabling the construction of appropriate control laws for real-time compensation of total disturbances. 
The preference for a linear ADRC approach is motivated by its practical advantages over nonlinear ADRC controllers \cite{c25}.
\subsection{Extended state space equation}
Taking the second-order derivative of the attitude dynamics \eqref{attitude
    dynamic} yields
\begin{equation}
    \label{second-order derivative of the attitude dynamics}
    \ddot{\eta} = \dot{\mathbf{Q}}\omega + \mathbf{Q}\dot{\omega}.
\end{equation}
By substituting \eqref{Newton-Euler equations} , we can obtain:
\begin{equation}
    \label{ddoteta}
    \ddot{\eta} = \mathbf{Q}\mathbf{J}^{-1}\mathbf{M} - \mathbf{Q}\mathbf{J}^{-1}\omega \times \mathbf{J}\omega + \dot{\mathbf{Q}}\omega.
\end{equation}
For controller design, we consider the effects of the last two terms in \eqref{ddoteta} as unknown disturbances and approximate $\mathbf{Q}$ as an identity matrix. 

Factors neglected during the modeling and control process, including nonlinear aerodynamic forces on the duct, disturbances during load attachment, model variations, and coupling of attitude dynamics, are collectively considered as the total disturbance $f$ of the system. 
Therefore \eqref{ddoteta} is simplified as:
\begin{equation}
    \label{simplified angular dynamics}
    \ddot \eta  = {\mathbf{J}^{ - 1}}\mathbf{M} + f .
\end{equation}

Taking the roll channel as an example, there are:

\begin{equation}
    \label{roll}
    \ddot \varphi  = \frac{1}{{{I_x}}}{\tau _x} + {f_x}.
\end{equation}

Where the disturbance ${f_x}$ is regarded as an augmented state, the extended
state vector $x = {\left[ {\begin{array}{*{20}{c}}
                    \varphi & {\dot \varphi } & {{f_x}}
                \end{array}} \right]^T}$  is taken, and the extended state space model can be written as:

\begin{equation}
    \label{state space model}
    \begin{array}{l}
        \dot x = \begin{bmatrix}
                     0 & 1 & 0 \\
                     0 & 0 & 1 \\
                     0 & 0 & 0
                 \end{bmatrix}x + \begin{bmatrix}
                                      0             \\
                                      \frac{1}{I_x} \\
                                      0
                                  \end{bmatrix}\tau_x + \begin{bmatrix}
                                                            0 \\
                                                            0 \\
                                                            1
                                                        \end{bmatrix}\dot{f}_x, \\
        y = \begin{bmatrix}
                1 & 0 & 0
            \end{bmatrix}x,
    \end{array}
\end{equation}
where $y = \varphi$ is the output of this system.

\begin{figure}
    \centering
    \includegraphics[width=\linewidth,]{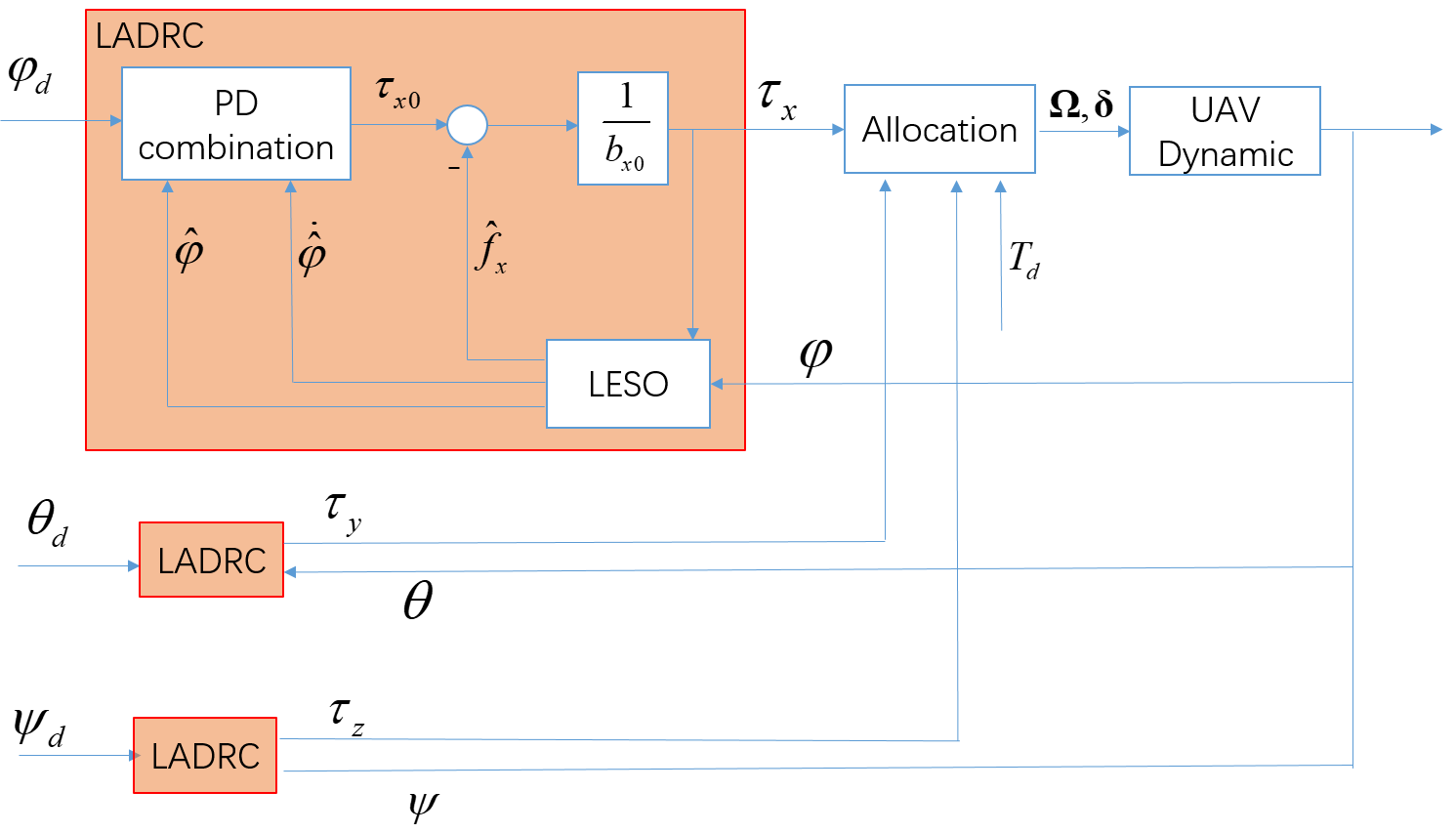}
    \caption{The structure of attitude controller based on LADRC}
    \label{fig:5}
\end{figure}

\subsection{Linear full order extended state observer}
To observe the state equation \eqref{state space model}, a state observer is
designed with the observer variable defined as $z = {\left[ {\begin{array}{*{20}{c}}
                    {\hat \varphi } & {\hat {\dot {\varphi}}} & {{{\hat f}_x}} 
                \end{array}} \right]^T}$ . 
The resulting observer is then obtained as:
\begin{equation}
    \label{Linear full order extended state observer}
    \begin{array}{l}
        \dot{z} = \mathbf{A}z + \mathbf{B}{\tau _x} + {\mathbf{L}_\varphi }(y - Cz), \\
        \hat{y} = \mathbf{C}z,
    \end{array}
\end{equation}

where

\[ \mathbf{A} = \begin{bmatrix}
        0 & 1 & 0 \\
        0 & 0 & 1 \\
        0 & 0 & 0
    \end{bmatrix}, \quad
    \mathbf{B} = \begin{bmatrix}
        0             \\
        \frac{1}{I_x} \\
        0
    \end{bmatrix}, \quad
    \mathbf{C} = \begin{bmatrix}
        1 & 0 & 0
    \end{bmatrix} \]
and ${\mathbf{L}_\varphi } = {\left[ {\begin{array}{*{20}{c}}
                    {{L_{\varphi 1}}} & {{L_{\varphi 2}}} & {{L_{\varphi 3}}}
                \end{array}} \right]^T}$ is the observer parameter. 
It is proved in \cite{c39} that, when the object model is known, the error of the ESO observer converges asymptotically to the origin. Even in the presence of significant model uncertainties, the estimation error of the observer remains bounded.

To enable the real-time operation of the ESO in flight control, it is discretized using the current Euler method described in \cite{c26}.

\subsection{Controller design}

Based on the observer, we can estimate the state of the system $\hat \varphi $
and $\hat {\dot {\varphi}}$, take them as feedback, and combine the desired
value of pitch angle ${\varphi _d}$ of the input signal to obtain reference
control quantity:
\begin{equation}
    \label{control law}
    {u_{x0}} = {k_{p\_roll}}\left( {{\varphi _d} - \hat \varphi } \right) - {k_{d\_roll}}\hat {\dot {\varphi}}.
\end{equation}
Here, ${k_{p\_roll}}$ and ${k_{d\_roll}}$ are control parameters. The reference control quantity obtained is subtracted from the
estimated disturbance ${\hat f_x}$ and divided by the scaling coefficient ${b_{x0}}$ to yield the system control quantity:
\begin{equation}
    \label{ux}
    {u_x} = \frac{1}{{{b_{x0}}}}\left( {{u_{x0}} - {{\hat f}_x}} \right).
\end{equation}
For pitch and yaw control, the control method mirrors that of the roll angle.
The complete attitude controller structure is illustrated in Fig. \ref{fig:5}.
\section{Experiment Results}
\subsection{Parameter Identification}
During flight, the attachment of payloads induces sustained torque disturbances, which need to be counteracted by deflecting the control vanes.
Therefore, knowing the magnitude of the torque that the control vanes can generate during flight is crucial. To this end, we constructed a test bench to acquire the unknown parameters in the model.
\begin{figure}[b]
    \centering
    \includegraphics[width=0.6\linewidth]{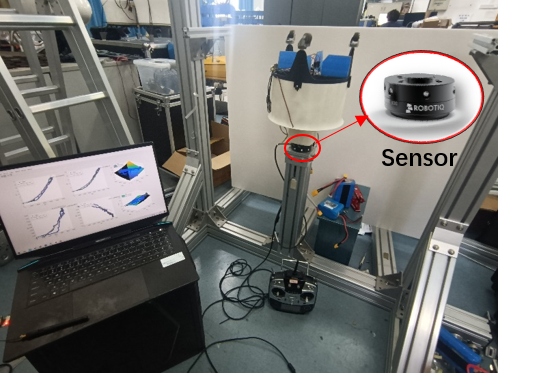}
    \caption{Test bench used for parameter identification. Invert the prototype and fix it onto the force and torque sensor.}
    \label{fig:6}
\end{figure}

As illustrated in Fig. \ref{fig:6}, the prototype is inverted and connected to a three-axis force and torque measurement module (Robotiq FT300). The remote controller provides reference motor speed and servo deflection angles. The following tests were conducted:

\subsubsection{Test 1: Control vane deflection torque}
\begin{figure}[t]
    \centering
    \includegraphics[width=\linewidth]{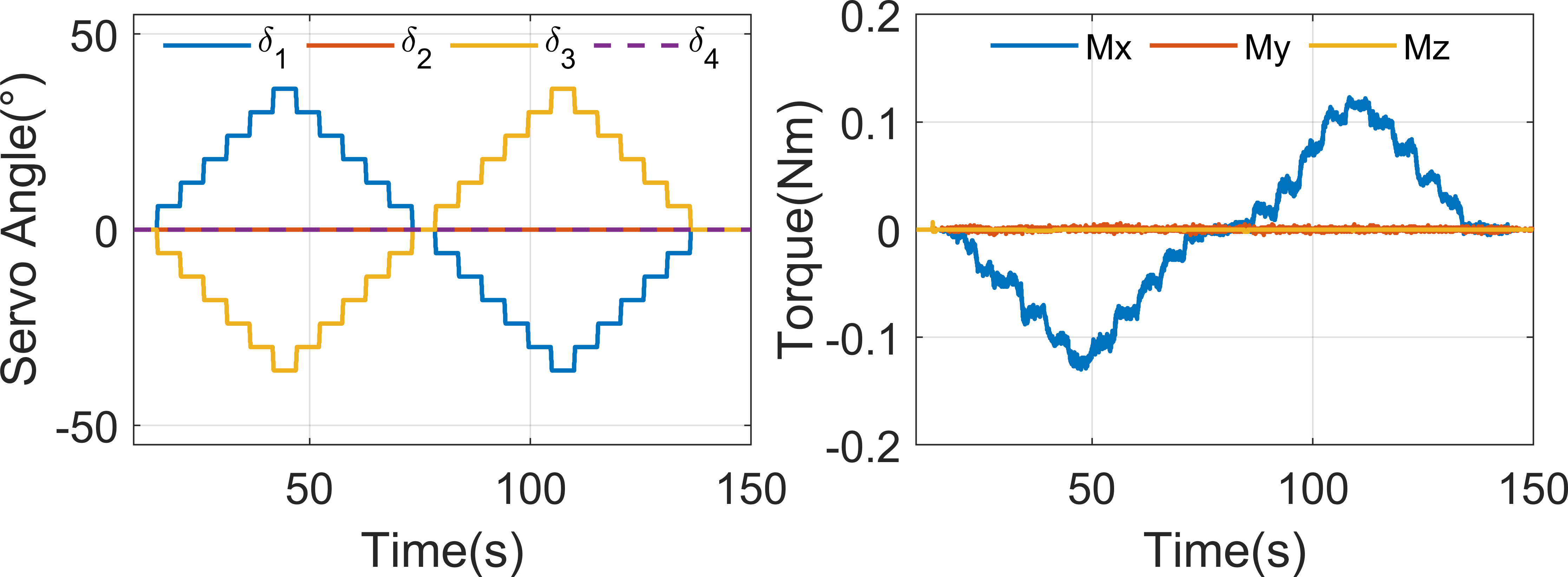}
    \caption{(a) Servo deflection signal variations during testing.(b) Control torque generated by vane rotation.}
    \label{fig:8}
\end{figure}

With two motors operating at the same rotational speed in near-hovering
conditions, a set of servos (e.g., servos 1 and 3) was systematically deflected
to their maximum angles determined by the mechanical structure, and the
resulting torque was measured.
As shown in Fig. \ref{fig:8}, utilizing collected servo angle and torque data to identify parameters based on the model \eqref{torque allocation}, yielding ${C_{{M_\delta }}} = 0.0014$. And the maximum moment of the control vanes is about 0.12Nm.

\begin{figure}[b]
    \centering
    {\includegraphics[width=\linewidth]{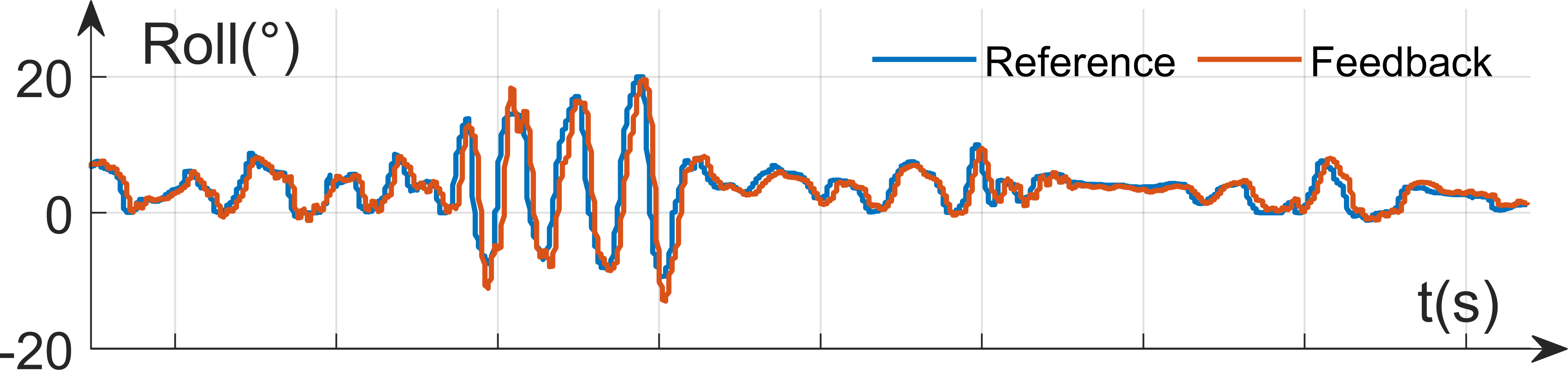}}\\
    {\includegraphics[width=\linewidth]{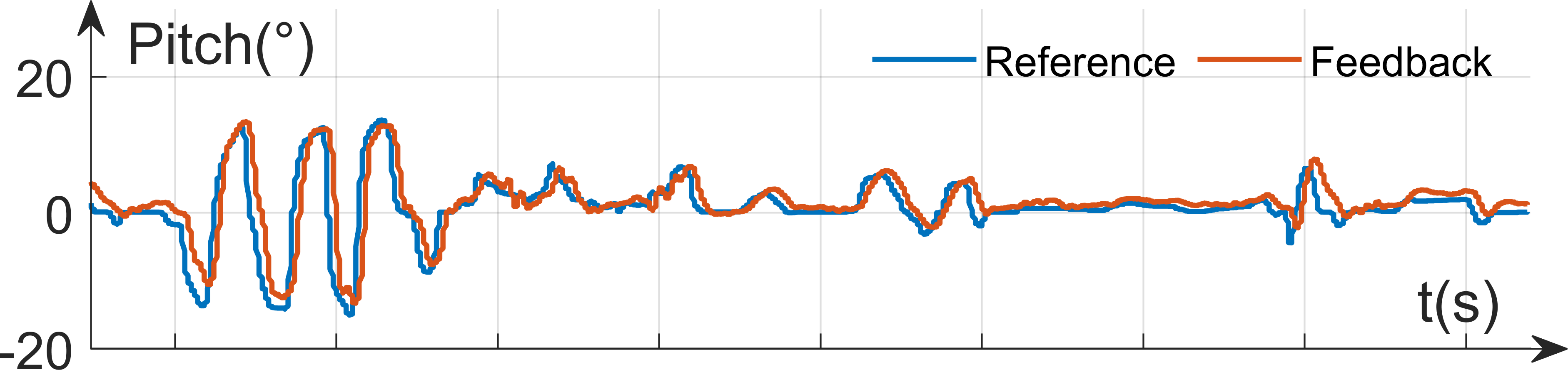}}\\
    {\includegraphics[width=\linewidth]{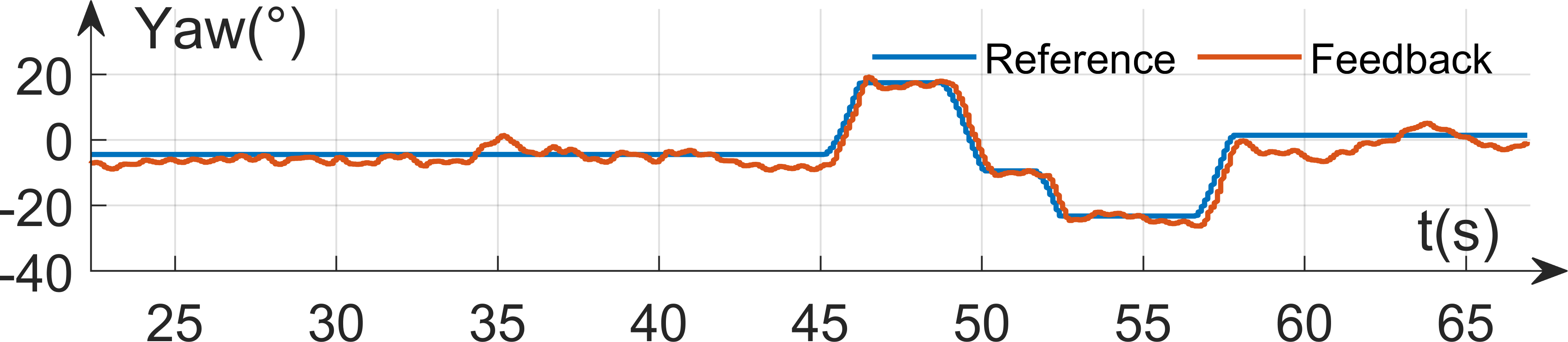}}\\
    \caption{The desired (blue line) and actual responses (red line) of the three-axis attitudes during flight without any loads.}
    \label{fig:free_attitude}
\end{figure}
\subsubsection{Test 2: Propeller thrust}
For the upper and lower motors, separate speed variations were applied, and
actual speed, Z-axis force, and Z-axis torque were simultaneously recorded.
The data was fitted to the model \eqref{DuctForce} \eqref{ductmoment} to obtain propeller parameters ${C_{Tz\_1}},{C_{Tz\_2}}$ and ${C_{Mz\_1}},{C_{Mz\_2}}$.

\subsection{Flight experiment}
To test the performance of the attitude controller proposed in this paper and verify the ability for safe aerial grabbing, we conducted the following experiments.
\begin{figure}[t]
    \centering
    {\includegraphics[width=\linewidth]{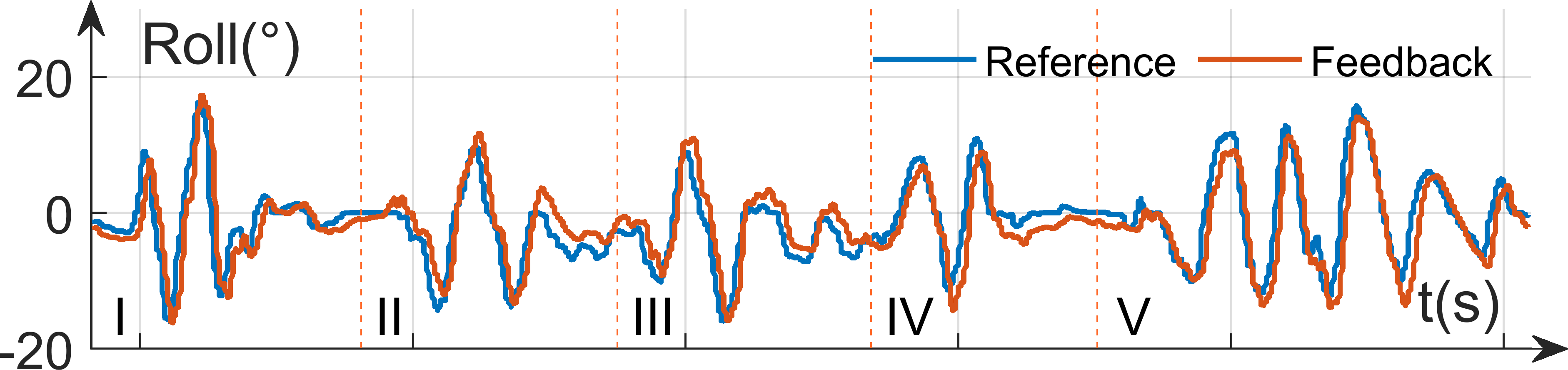}}\\
    {\includegraphics[width=\linewidth]{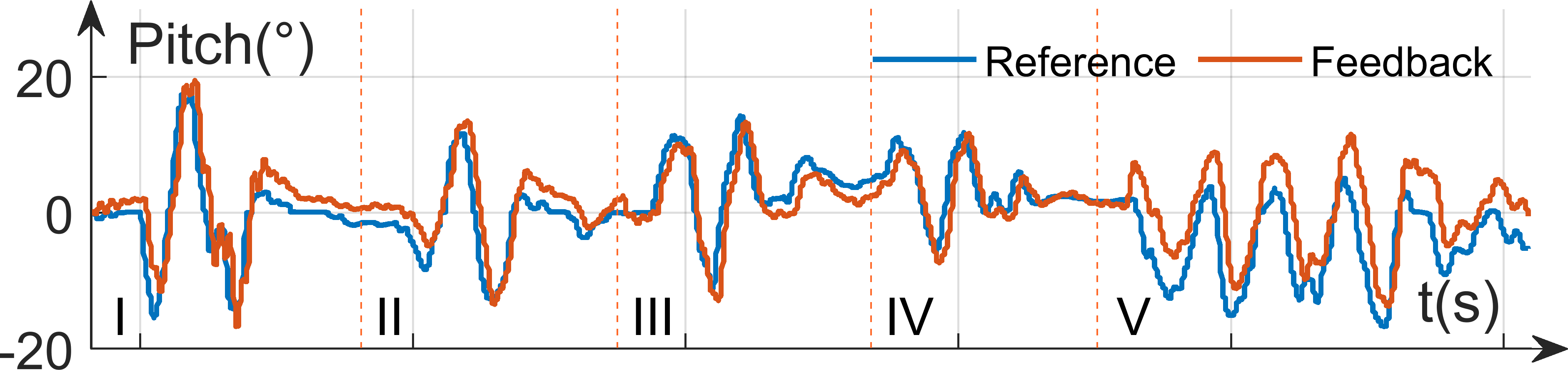}}\\
    {\includegraphics[width=0.98\linewidth]{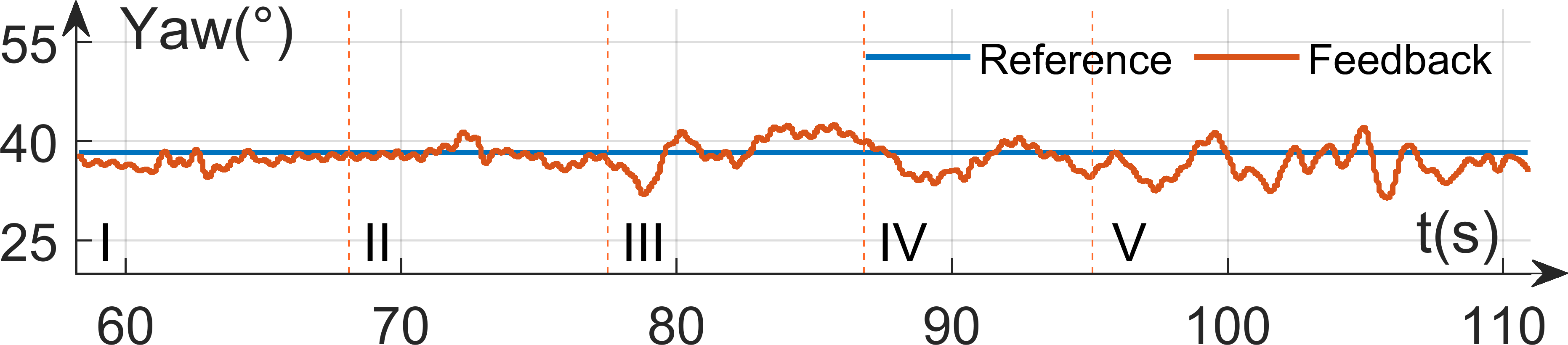}}\\
    \caption{The desired (blue line) and actual responses (red line) of the three-axis attitudes during load attachment.}
    \label{grab_attitude}
\end{figure}
\begin{figure}[t]
    \centering
    \begin{subfigure}[b]{0.5\columnwidth}
        \centering
        \includegraphics[height=1.1in]{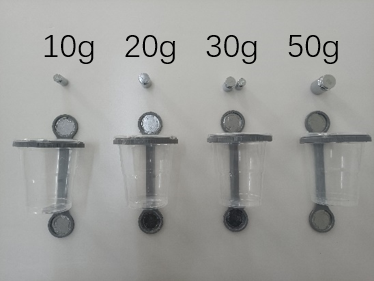}
        \hspace{0.001\columnwidth} %
        \caption{}
        \label{fig:10a}
    \end{subfigure}
    \begin{subfigure}[b]{0.4\columnwidth}
        \centering
        \includegraphics[height=1.1in]{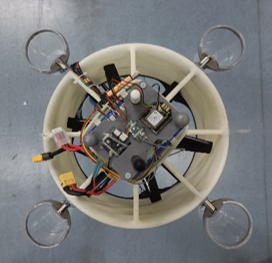}
        \hspace{0.001\columnwidth} %
        \caption{}
        \label{fig:10b}
    \end{subfigure}
    \caption{
        (a) Using weights of different masses to represent various loads
        (b)Top view of the UAV with all mounting points loaded.
    }
    \label{fig:10}
\end{figure}
\subsubsection{Test 1: Attitude tracking test}
After the aircraft takes off, a reference attitude signal is provided via
remote control, and Fig. \ref{fig:free_attitude} records the three-axis attitude
of the aircraft. The blue line in the figure represents the desired attitude,
while the red line depicts the sensor feedback. It can be seen that despite
frequent changes in the reference attitude, the controller effectively
maintains the attitude error within a small range.
 
\subsubsection{Test 2: Safe Aerial Grabbing}
The designed aircraft aims to safely achieve multiple payload attachments and transportation in mid-air. To evaluate this capability, individuals approached the drone during flight and manually loaded various weights onto the electromagnet attaching points. 
Throughout this process, the reference attitude signal was given by remote controller. 
Weights (10g, 20g, 30g, and 50g) were mounted in counterclockwise order on the electromagnets, while the mass of the module used for load placement is 15g(as shown in Fig. \ref{fig:10}).
\begin{figure}[h]
    \centering
    \subcaptionbox{}{\includegraphics[width=1.4in,height=0.3\linewidth, angle=-90]{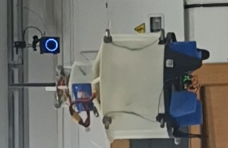}}
    \subcaptionbox{}{\includegraphics[width=1.4in,height=0.3\linewidth, angle=-90]{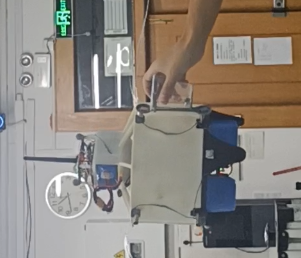}}
    \subcaptionbox{}{\includegraphics[width=1.4in,height=0.3\linewidth, angle=-90]{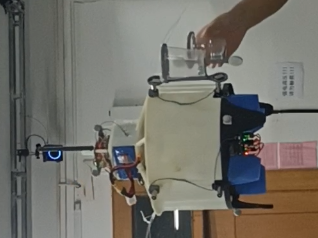}}\\
    \subcaptionbox{}{\includegraphics[width=1.4in,height=0.3\linewidth, angle=-90]{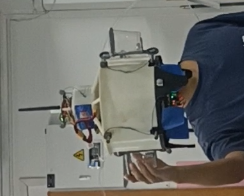}}
    \subcaptionbox{}{\includegraphics[width=1.4in,height=0.3\linewidth, angle=-90]{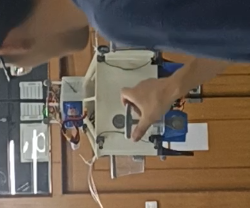}}
    \subcaptionbox{}{\includegraphics[width=1.4in,height=0.3\linewidth, angle=-90]{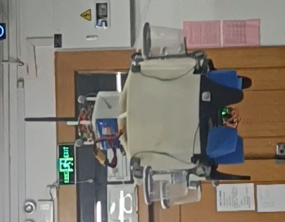}}
    \caption{Safe aerial grabbing experiment: Repeated contact with UAV and placement of loads during flight. (a)Free flight without any loads.(b)First attachment with a 25g load. (c)Second attachment with a 35g load. (d)Third attachment with a 45g load. (e)Fourth attachment with a 65g load. (f)Flight with all mounting points loaded}
    \label{fig:12 }
\end{figure}
The process involves maneuver flight after takeoff, attitude adjustment to hover, and load attachment, followed by another maneuver flight to test attitude tracking. This sequence is repeated until all attachment points carry loads, and finally, attitude tracking under full loads is evaluated.
Fig. \ref{grab_attitude} illustrates the aircraft's attitude during the gradual attachment of different weights. 
(Roman capital letters in the figure represent different stages of flight, where I is free flight without any loads, and an additional load is added at each stage from II to V.)
The controller effectively resists the disturbance caused by the load attachment, keeping the error within an acceptable range during the overall attitude tracking process. 
The specific mounting process is depicted in Fig. \ref{fig:12 }. 
We recommend the readers refer to the attached video for more details of this experiment.

\section{Conclusion and future research}
This paper presents the design, implementation, modeling, and controller design of a novel coaxial ducted fan UAV called Ductopus. Multiple payloads can be safely and straightforwardly attached to the UAV during flight, which provides a good choice for aerial grasping and other aerial manipulation tasks. 
Through experiments, it has been demonstrated that the selected LADRC controller can resist disturbances from the payload and effectively achieve attitude tracking.
In the current prototype design, the use of 3D printing materials has resulted in excessive weight of the airframe, which may lead to overloading of the motor after too many payloads are added. Subsequent designs will reduce the structural weight of the aircraft to improve its usability. Looking ahead, future research will focus on enhancing position control for increased practicality. The integration of modular payloads, allowing mounting points for diverse functionalities, will be pursued to broaden the UAV's range of applications.

\bibliographystyle{ieeetr}
\bibliography{reference}

\end{document}